\documentclass{article}
\usepackage{ijcai20}
\usepackage{bbm}
\usepackage{times}

\usepackage{soul}
\usepackage{url}

\usepackage[hidelinks]{hyperref}
\usepackage[utf8]{inputenc}
\usepackage[small]{caption}
\usepackage{graphicx}
\usepackage{makecell}
\usepackage{subcaption}
\usepackage{amsmath}
\usepackage{booktabs}
\title{Learning from Noisy Similar and Dissimilar Data}

\author{
Soham Dan$^1$\footnote{Work done during an internship at RIKEN}\and
Han Bao$^{2,3}$\and
Masashi Sugiyama$^{2,3}$\\
\affiliations
$^1$University of Pennsylvania\\
$^2$The University of Tokyo\\
$^3$RIKEN Center for Advanced Intelligence Project\\
\emails
sohamdan@seas.upenn.edu,
tsutsumi@ms.k.u-tokyo.ac.jp,
sugi@k.u-tokyo.ac.jp
}

\begin{document}

\maketitle

\begin{abstract}
With the widespread use of machine learning for classification, it becomes increasingly important to be able to use weaker kinds of supervision for tasks in which it is hard to obtain standard labeled data. One such kind of supervision is provided pairwise---in the form of Similar (S) pairs (if two examples belong to the same class) and Dissimilar (D) pairs (if two examples belong to different classes). This kind of supervision is realistic in privacy-sensitive domains. Although this problem has been looked at recently, it is unclear how to learn from such supervision under label noise, which is very common when the supervision is crowd-sourced. In this paper, we close this gap and demonstrate how to learn a classifier from noisy S and D labeled data. We perform a detailed investigation of this problem under two realistic noise models and propose two algorithms to learn from noisy S-D data. We also show important connections between learning from such pairwise supervision data and learning from ordinary class-labeled data. Finally, we perform experiments on synthetic and real world datasets and show our noise-informed algorithms outperform noise-blind baselines in learning from noisy pairwise data.

\end{abstract}

\section{Introduction}

In supervised classification, a classifier is trained with labeled training data points, which are usually collected through human annotation.
While collecting labeled data points is the traditional way to apply supervised classification, \emph{pairwise comparison} is often more appealing for human decision making~\cite{furnkranz2010preference}, where annotators are requested to compare two instances and give relative relationships between them;
e.g., which instance has stronger stimulus, whether two instances belong to the same category, and so on.
This is partly because (1) decision makers tend to be subjective at directly choosing a single hypothesis\footnote{\cite{thurstone1927law} has studied a relationship between relative comparison and a single hypothesis on stimuli, which is known as the law of comparative judgement.},
and (2) decision makers are often biased about picking an opinion.\footnote{This bias is known as social desirability bias~\cite{fisher1993social}; questionees are unconsciously led to a socially desirable opinion when they are asked to reveal their opinions in a direct way. Such a tendency is observed especially in answering their sensitive matters such as criminal records.}
Thus, pairwise comparison has been exploited in a number of applications such as pairwise ranking~\cite{furnkranz2010preference,jamieson2011active}, Bayesian optimization~\cite{eric2008active}, derivative-free optimization~\cite{jamieson2012query}, and analytic hierarchical processes~\cite{saaty1990decision}.
\\
One of the methods that incorporate pairwise comparison into identification of latent categories of data is \emph{semi-supervised clustering}~\cite{basu2008constrained},
which utilizes pairwise supervision indicating whether two instances belong to the same cluster or not (known as must-link and cannot-link constraints), guiding clustering as decision makers desire.
Recently, \cite{bao2018classification} and \cite{shimada2019classification} gave an empirical risk minimization (ERM) formulation to train an inductive classifier from pairwise comparison,
which has successfully connected supervised learning and pairwise comparison and avoided dataset-dependent assumptions used in semi-supervised clustering.

However, there is one important gap that has not been addressed in classification with pairwise supervision---real world data collection is bound to be noisy. Several previous works address noise in ordinary supervision with class-labeled data---\cite{natarajan2013learning,patrini2017making,jiang2017mentornet,han2018co}.
For pairwise supervision, there are two unique types of error.
The first error results from \emph{pairing corruption}: some pairs of instances are hard to identify whether they belong to the same category or not.
The second error is from \emph{labeling corruption}: labels of some instances are intrinsically ambiguous thus subsequent pairwise comparison is also affected.
Depending on these situations, we can consider two types of noise models in pairwise supervision.

In this paper, we investigate classification with noisy pairwise supervision, where noise is present in pairwise comparison and follows either pairing corruption or labeling corruption.
Our first strategy is inspired by previous work \cite{patrini2017making,natarajan2013learning} that uses standard class-labeled data with class-conditional noise to learn a classifier.
We introduce a corrected loss function, which induces an unbiased estimator of the classification risk in the presence of noise for pairwise data.
Subsequently, a classifier can be obtained through the minimization of the corrected loss.
The second strategy is based on  weighted empirical risk minimization, or  cost-sensitive classification \cite{elkan2001foundations,scott2012calibrated}.
This is motivated from the insight that the Bayes classifier of the classification risk under the noise-free distribution corresponds to that of the weighted risk under the noisy distribution.

To the best of our knowledge, this is the first work to investigate learning from noisy pairwise data and the following are our main contributions :

\begin{itemize}
\item Definition of realistic noise models for the S-D learning problem.
\item Providing two algorithms based on loss correction and weighted classification that can effectively learn from noisy S-D training data (under either noise model) and still give good performance (standard binary classification accuracy) on clean test data.
\item Proving that the Bayes classifier for noise-free and symmetric-noise S-D learning is identical to the Bayes classifier for standard binary classification.
\item We analyze the generalization bounds of our algorithms and provide two new generalization bounds for clean S-D learning \cite{shimada2019classification}
\item Performing experiments on various datasets to show that the proposed algorithms work well in practice.
\end{itemize}

\section{Problem Setup}
Let $\mathcal{X}$ denote the instance space, $\mathcal{Y}$ the label space and $\mathcal{Z}$ the
the underlying distribution over $(\mathcal{X},\mathcal{Y})$. $\mathcal{Z}$ is the distribution with respect to which we want to perform well---the test data is drawn from this distribution and the performance metric is classification accuracy : $\sum_{(x,y)\in test_data}\mathbbm{1}[y_{predicted} \neq y]$, where $\mathbbm{1}[\cdot]$ denotes the indicator function. However, we assume that due to the domain constraints we are unable to procure direct class-labeled data from $\mathcal{Z}$ and only have access to pairwise supervision---whether a pair of instances $(X,X')$ is from the same class ($Y=Y'$) or from different classes ($Y \neq Y'$). There is a latent variable $Q$ dictating whether the sample is drawn from similar data ($Q=1$) or it is drawn from dissimilar data ($Q=-1$). Let $\mathcal{Z}_Q$ denote the distribution over $(\mathcal{X},Q)$. Further let $\pi=P(Y=+1)$ denote the class prior and $P_+(X)=P(X|Y=+1)$ and $P_-(X)=P(X|Y=-1)$. Now we present the two noise models that reflect what we may expect in real-world data.  

\begin{figure}
    \centering
    \includegraphics[width=\linewidth]{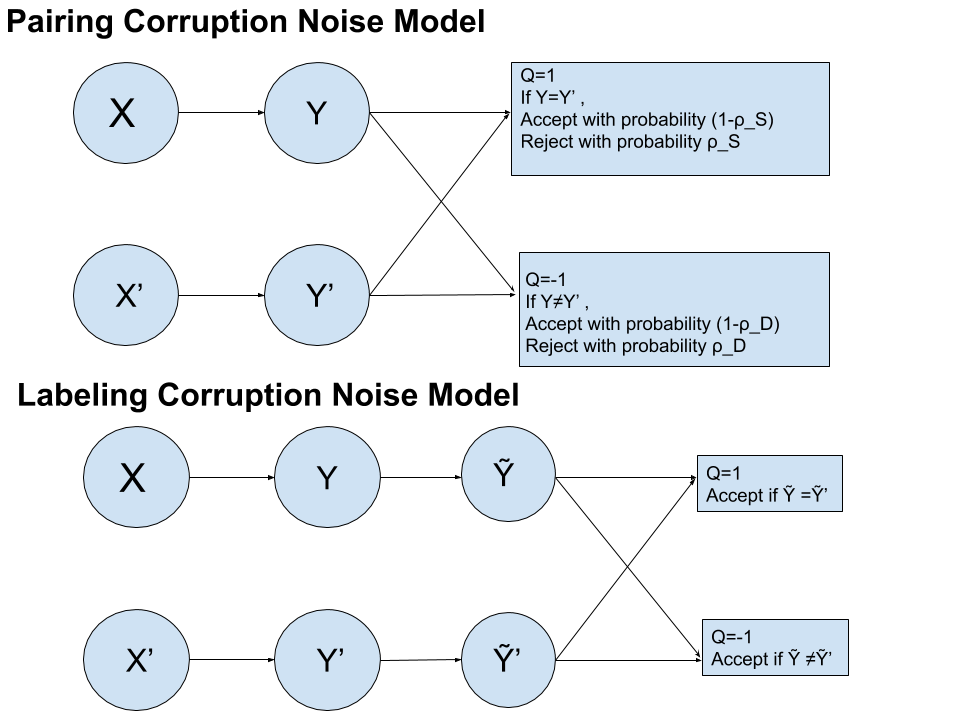}
    \caption{The two noise models}
    \label{fig:my_label}
\end{figure}

\subsection{Noise Model 1 : Pairing Corruption}
This model is motivated by the following scenario--- imagine non-perfect crowd-workers who are given a pool of instances $X_i$ and annotate pairs $(X_i,X_j)$ as $Q=1$, if they believe $Y_i=Y_j$ and $Q=-1$ otherwise. Since they are non-perfect, they make mistakes in this process and sometimes assigns $Q=-1$ when it should have been $Q=1$ and vice versa. Thus, in this model two samples $(X,Y)$ and $(X',Y')$ are drawn from the underlying distribution $\mathcal{Z}$ and if $Y=Y'$, this pair of sample is classified as Similar ($Q=1$) with probability $1-\rho_S$ and if $Y \neq Y'$ this pair is classified as Dissimilar ($Q=-1$) with probability $1-\rho_D$. In other words, the noise rate for similar (S) and dissimilar (D) data are $\rho_S$ and $\rho_D$ respectively. Thus, under this noise model, each of the S and D samples are drawn from mixtures of the true S and D distributions :  $P(Q=1|Y=Y')=1-\rho_S$ and $P(Q=-1|Y \neq Y')=1-\rho_D$. This noise model is instance-independent but label-dependent (different noise rates for $Q=1$ and $Q=-1$).

\subsection{Noise Model 2: Labeling Corruption}
Consider that we are dealing with a privacy sensitive domain and responders do not want their individual labels sent to the learning module---further, some of them lie about their labels so the labels are noisy. A moderator converts the point-wise data $(X,Y)$ to pairwise data $(X,X',Q= \pm 1)$, to preserve privacy. Mathematically, we first generate two samples $(X,Y)$ and $(X',Y')$ from the underlying distribution $D$ and then the label is flipped with probability $\rho_y$ (this is class conditioned : if a sample originally has label $+1$ it is flipped to label $-1$ with probability $\rho_{+}$ and respectively, $\rho_{-}$ for the other case). $P(\tilde{Y}=+1|Y=-1)=\rho_{-}$ and $P(\tilde{Y}=-1|Y=+1)=\rho_{+}$ Then in the next step the similar or dissimilar labels are assigned (there is assumed to be no noise in this step since the moderator is an expert). Thus, $P(Q=1|\tilde{Y}=\tilde{Y'})=1$ and $P(Q=-1|\tilde{Y} \neq \tilde{Y'})=1$.

If there were no noise corruption, the S data would comprise of two positive instances or two negative instances and the $D$ data would comprise of one positive instance and one negative instance. Accordingly, in the noise-free pairwise data scenario the densities ($P_S(x,x')$,$P_D(x,x')$) of the S and D data are respectively :
\begin{equation} \label{eq1}
\begin{split}
P_S(x,x') & = \frac{\pi^2 P_+(x)P_+(x')+(1-\pi)^2P_-(x)P_-(x')}{\pi^2+(1-\pi)^2},
\\
P_D(x,x') & = \frac{\pi (1-\pi) P_+(x)P_-(x')+\pi(1-\pi)P_-(x)P_+(x')}{2\pi(1-\pi)}.
\end{split}
\end{equation}
Further, one can marginalize out $x'$ and get the densities in terms of a single $x$ being drawn from either $S$ or $D$. This view is important for our analysis and has been previously used in \cite{bao2018classification}. The implication is that we can now treat the S-D data as point-wise data from the Similar (S) and Dissimilar (D) classes.
\begin{equation} \label{eq2}
\begin{split}
P_S(x) & = \frac{\pi^2 P_+(x)+(1-\pi)^2P_-(x)}{\pi^2+(1-\pi)^2},
\\
P_D(x) & = \frac{ P_+(x)+P_-(x)}{2}.
\end{split}
\end{equation}
The above expressions are derived in \cite{bao2018classification} under the noise-free assumption. Here we present the expression for the densities ($\tilde{P}_S(x)$---similar expressions can be obtained for $\tilde{P}_D(x)$)  under each of the noise models presented above. For the pairing corruption model we get :
\begin{equation} \label{eq2}
\tilde{P}_S(x) = (1-\rho_S)P_S(x) + \rho_DP_D(x) ,
\end{equation}
and for the labeling corruption model we get, $\tilde{P}_S(x)$
\begin{equation} \label{eq2}
\begin{split}
& = \frac{(\pi(1-\rho_{+})P_+(x)+(1-\pi)\rho_{-}P_-(x)) \Tilde{\pi}}{\Tilde{\pi}^2+(1-\Tilde{\pi})^2},
\\
& +\frac{((1-\pi)(1-\rho_{-})P_-(x)+\pi\rho_{+}P_+(x))(1-\Tilde{\pi})} {\Tilde{\pi}^2+(1-\Tilde{\pi})^2},
\\
& \text{where} \quad \Tilde{\pi}=(\pi(1-\rho_{+})+(1-\pi)\rho_{-}).
\end{split}
\end{equation}

\begin{figure*}[!h]
    \centering 
\begin{subfigure}{0.18\textwidth}
  \includegraphics[width=\linewidth]{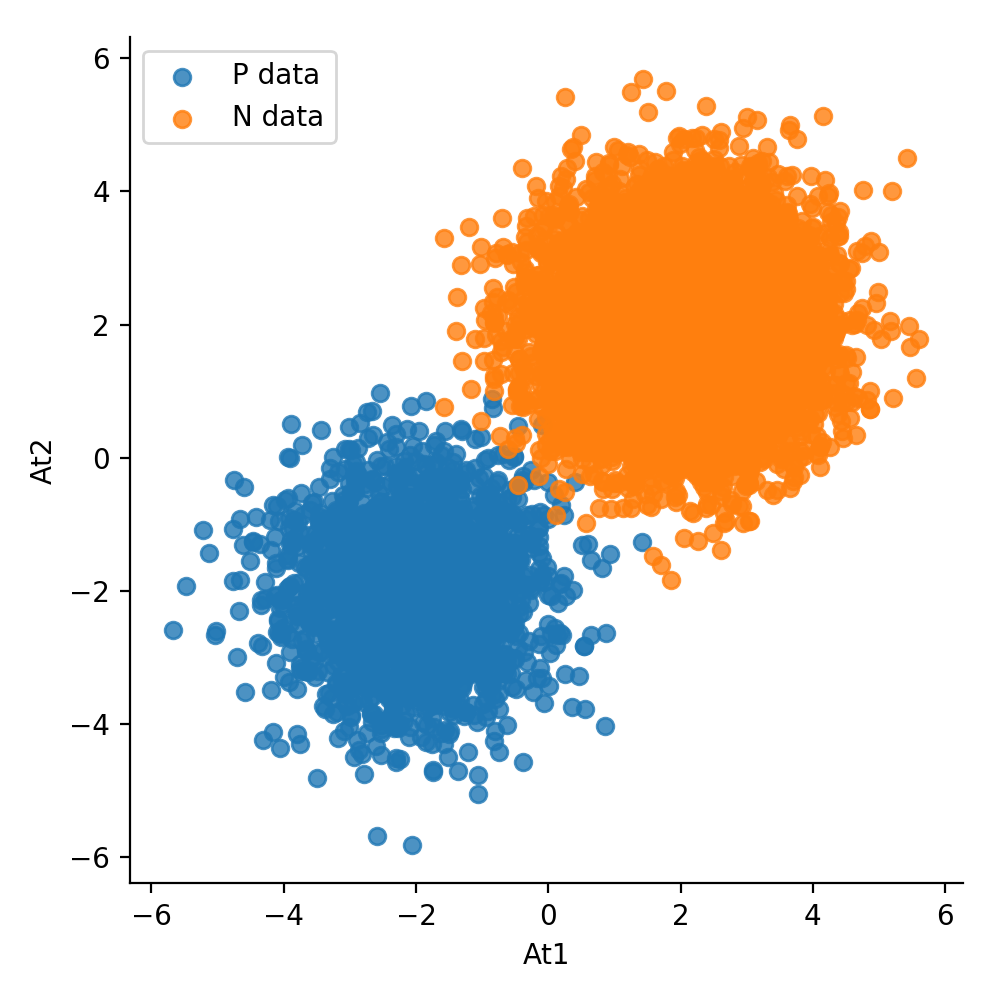}
  \caption{Underlying P-N distribution $D$}
  \label{fig:1}
\end{subfigure}\hfil 
\begin{subfigure}{0.18\textwidth}
  \includegraphics[width=\linewidth]{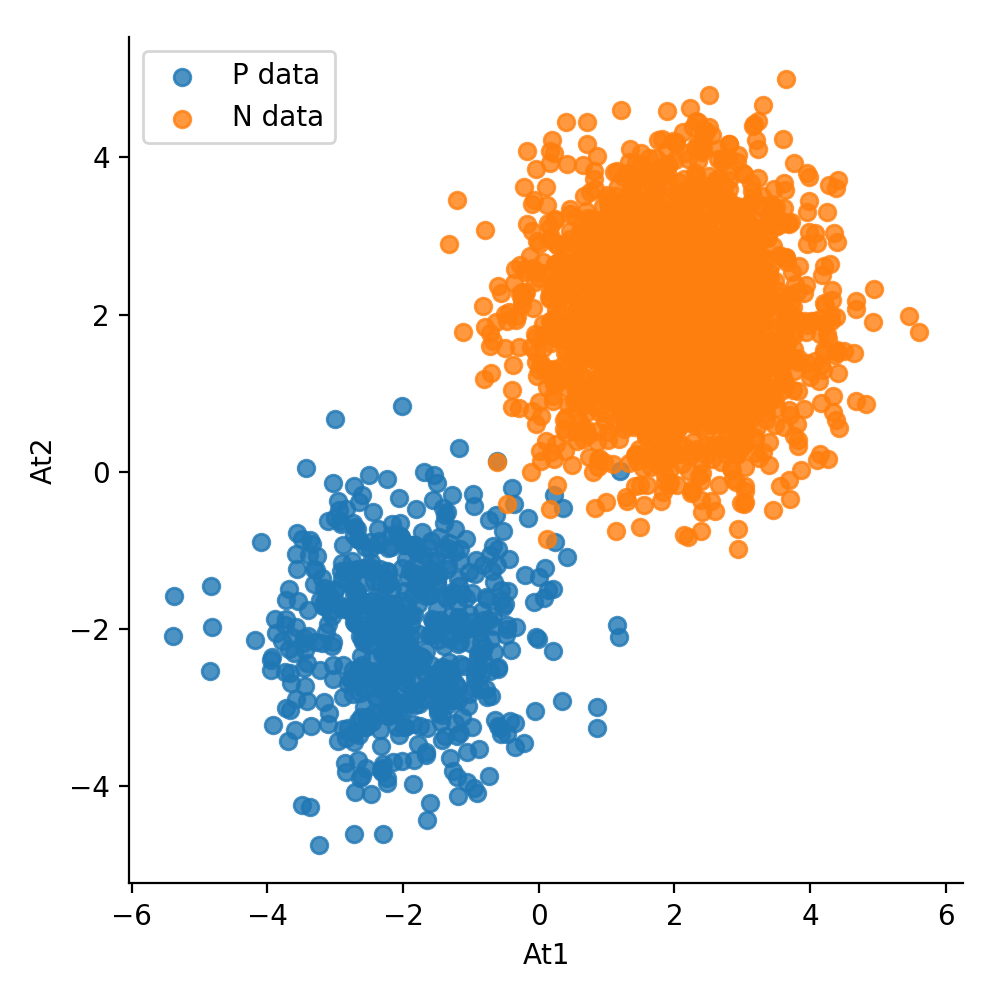}
  \caption{Test Set of $3000$ points}
  \label{fig:2}
\end{subfigure}\hfil 
\begin{subfigure}{0.18\textwidth}
  \includegraphics[width=\linewidth]{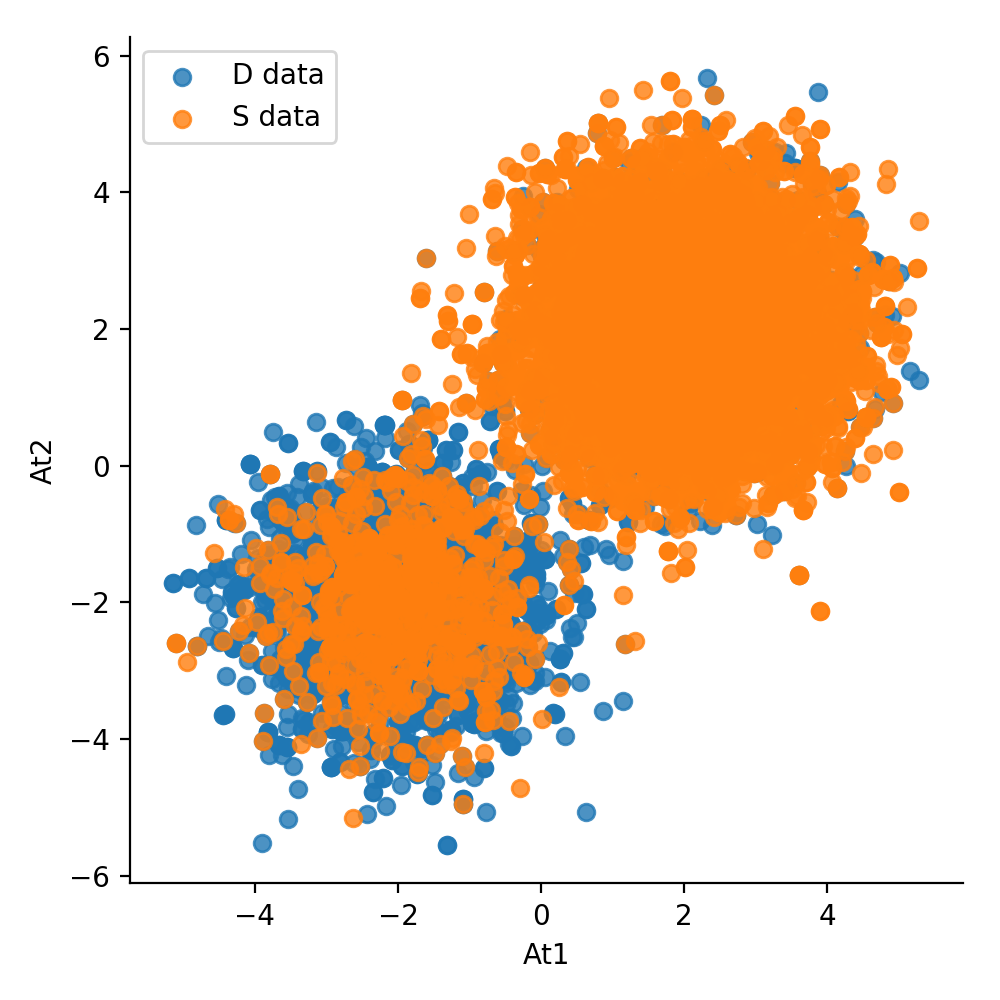}
  \caption{S and D points ($20,000$) }
  \label{fig:3}
\end{subfigure}
\begin{subfigure}{0.18\textwidth}
  \includegraphics[width=\linewidth]{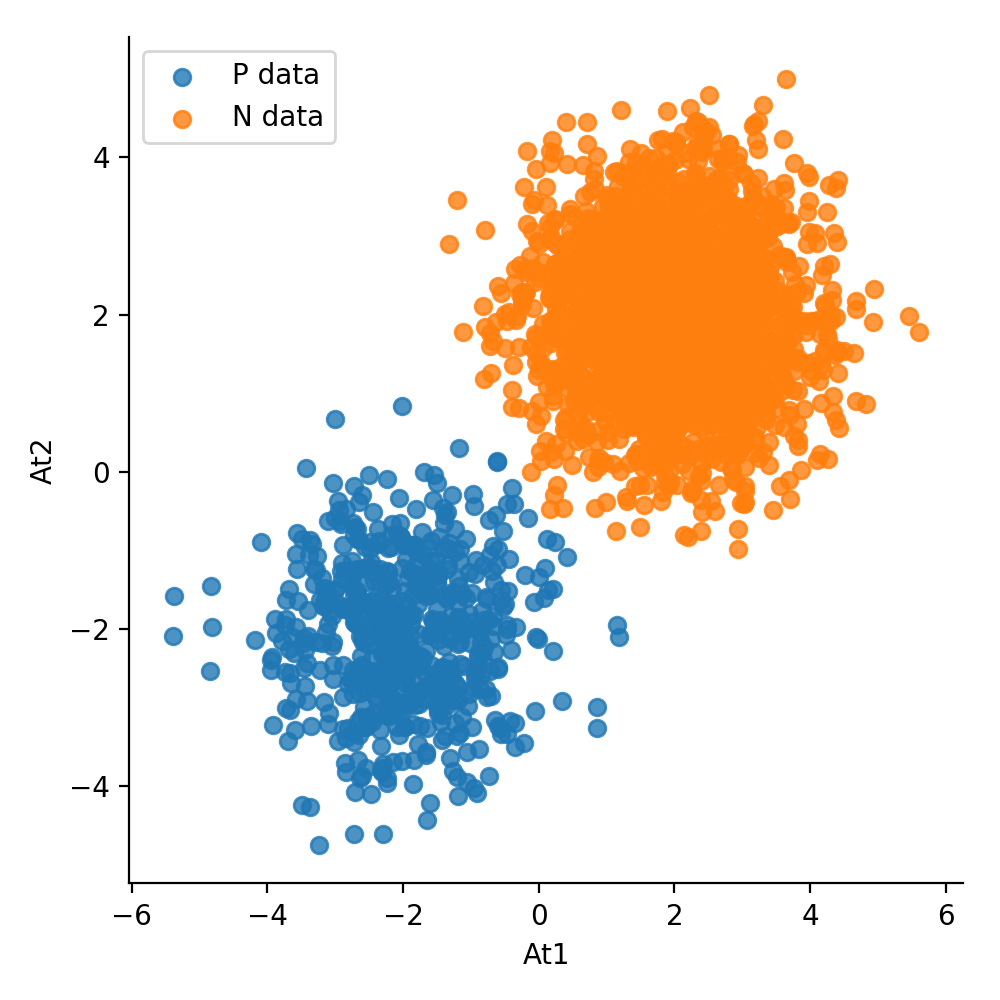}
  \caption{Weighted Classification: $99.8\%$ accuracy}
  \label{fig:2}
\end{subfigure}\hfil 
\begin{subfigure}{0.18\textwidth}
  \includegraphics[width=\linewidth]{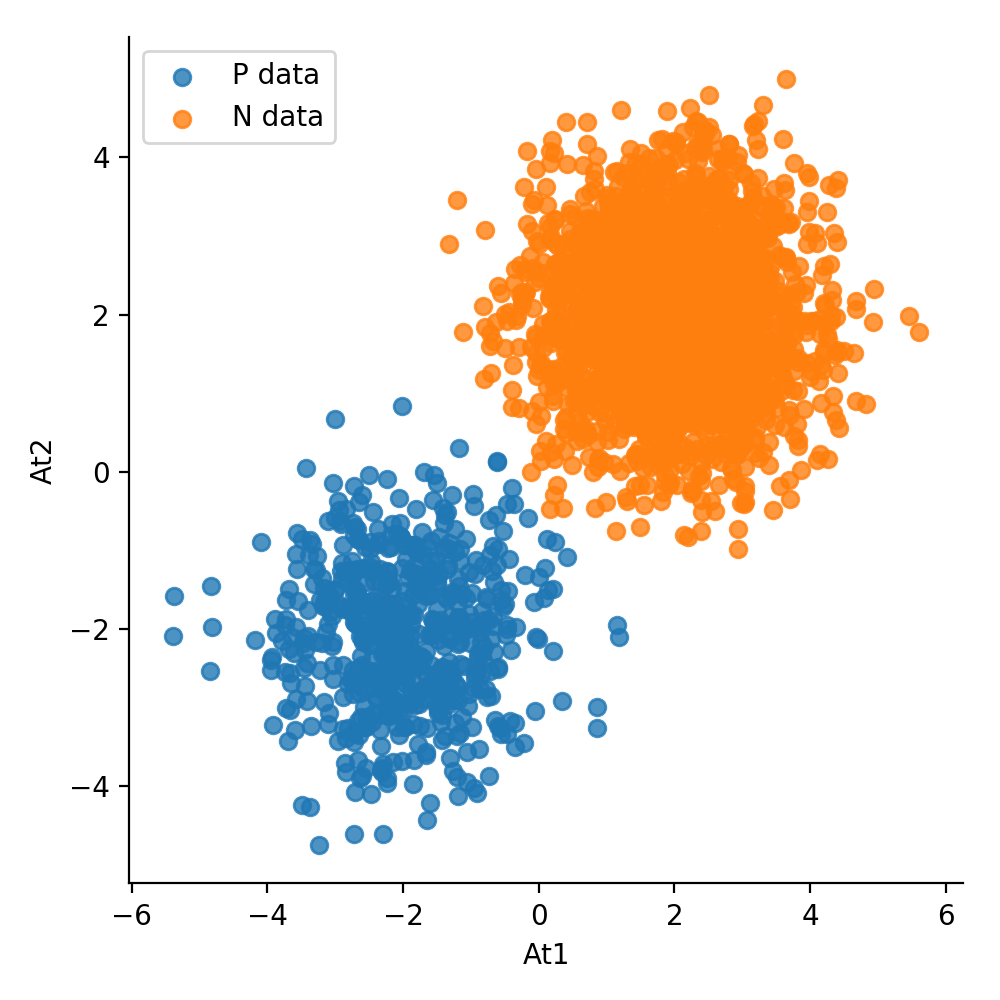}
  \caption{Loss Correction: $99.8\%$ accuracy}
  \label{fig:3}
\end{subfigure}
\medskip
\begin{subfigure}{0.15\textwidth}
  \includegraphics[width=\linewidth]{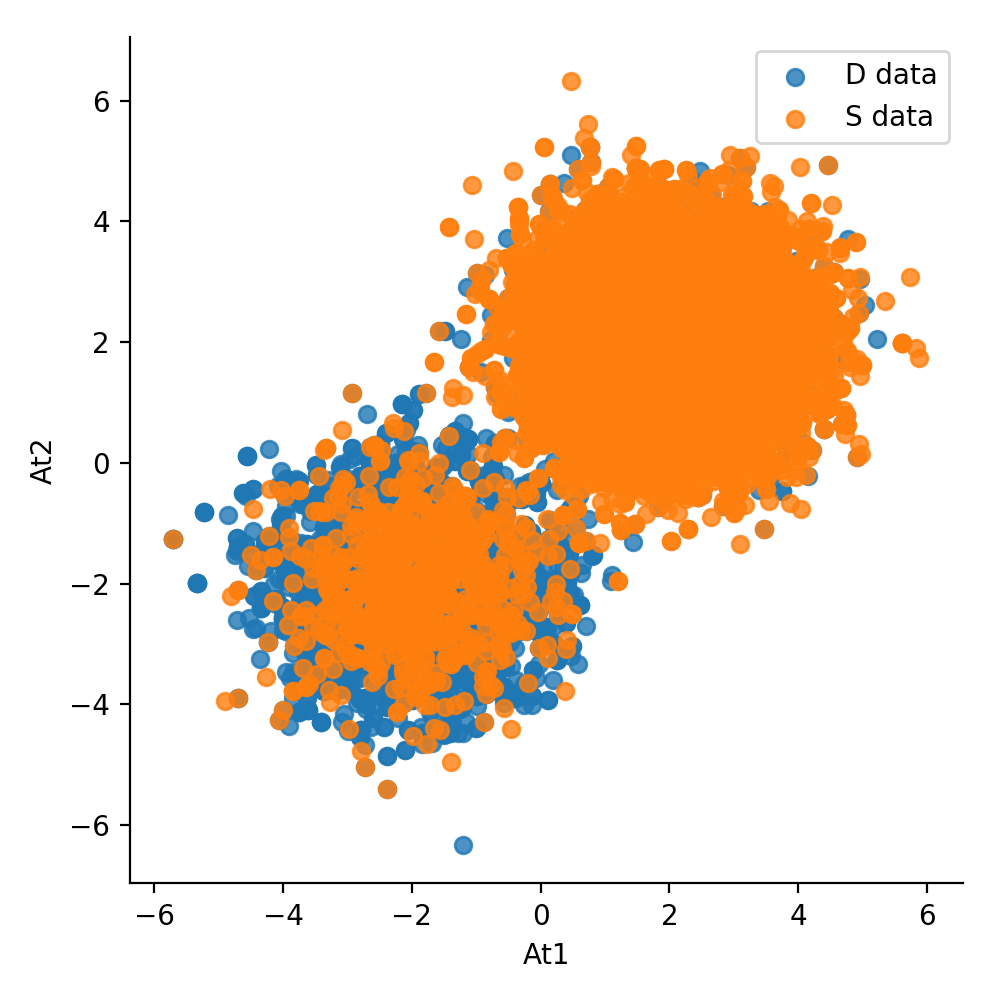}
  \caption{noisy S-D data (Pairing Noise)}
  \label{fig:4}
\end{subfigure}\hfil 
\begin{subfigure}{0.15\textwidth}
  \includegraphics[width=\linewidth]{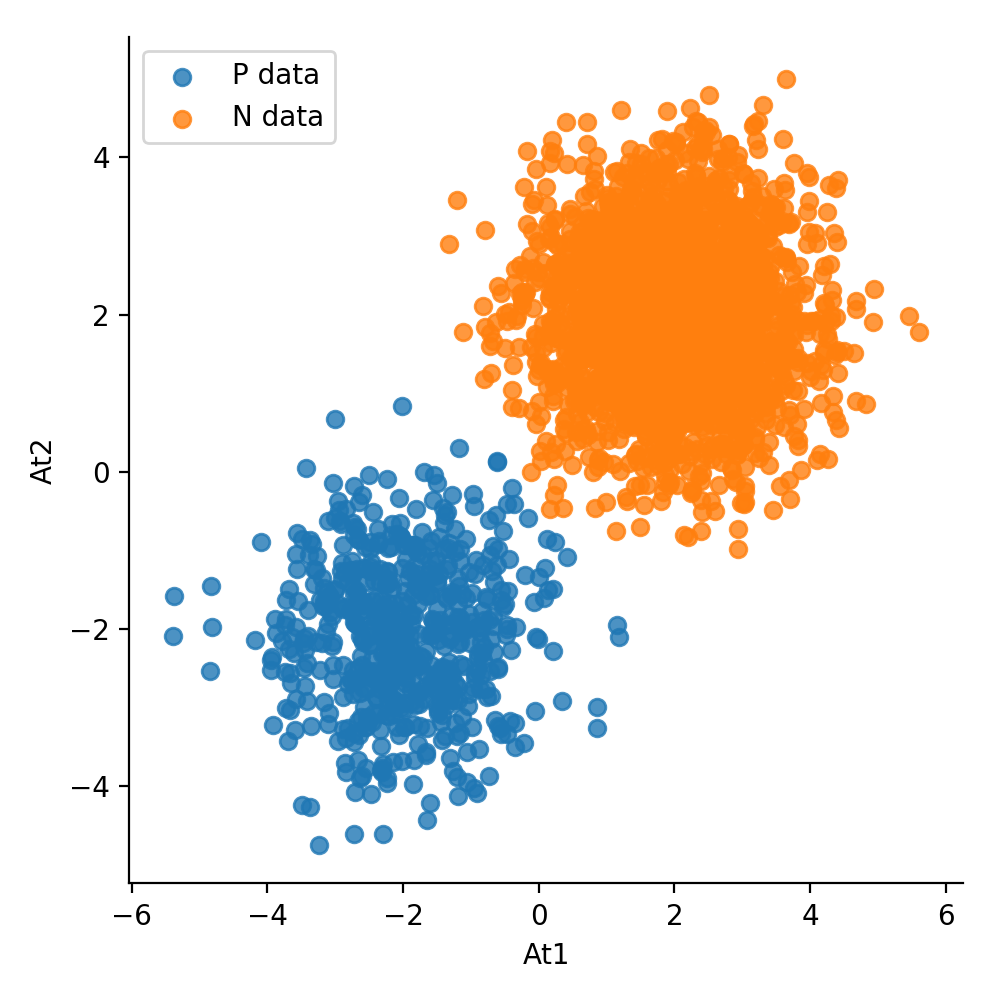}
  \caption{Weighted Classification: $99.73\%$ accuracy}
  \label{fig:5}
\end{subfigure}\hfil 
\begin{subfigure}{0.15\textwidth}
  \includegraphics[width=\linewidth]{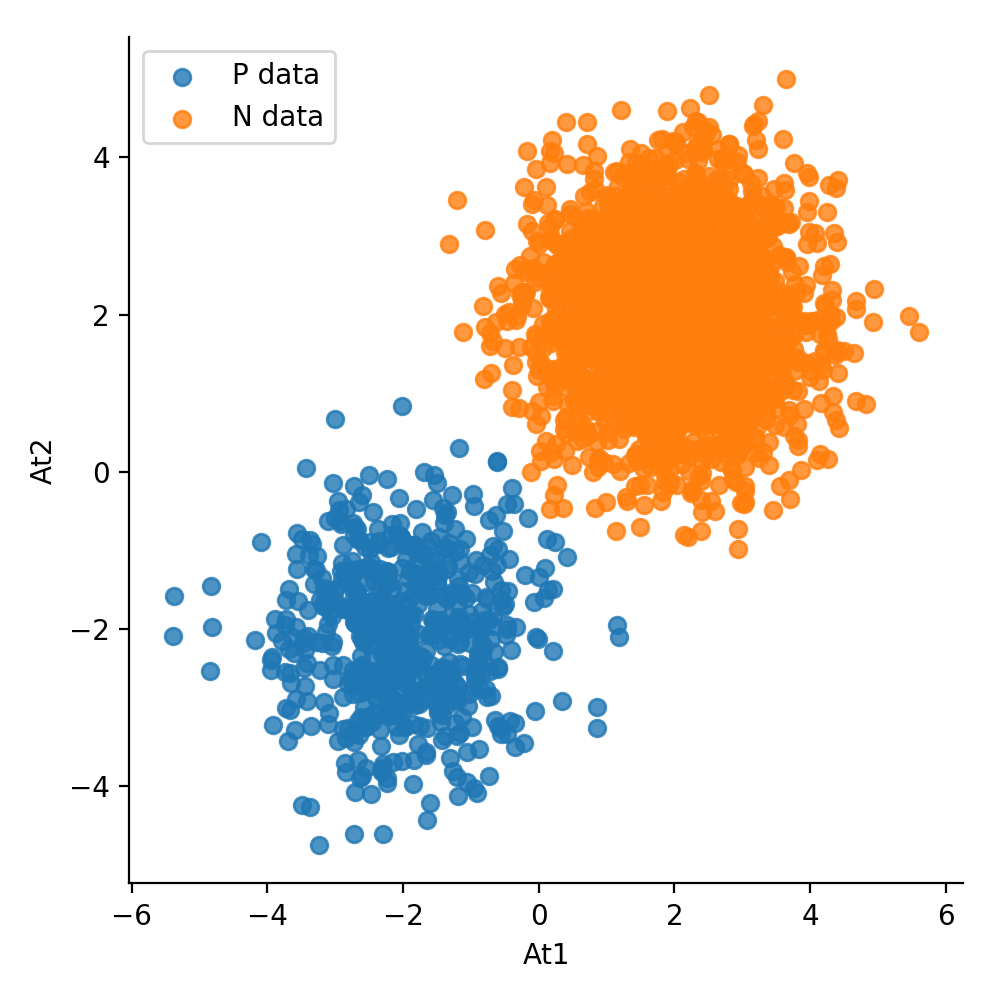}
  \caption{Loss Correction: $99.73\%$ accuracy}
  \label{fig:6}
\end{subfigure}\hfil 
\begin{subfigure}{0.15\textwidth}
  \includegraphics[width=\linewidth]{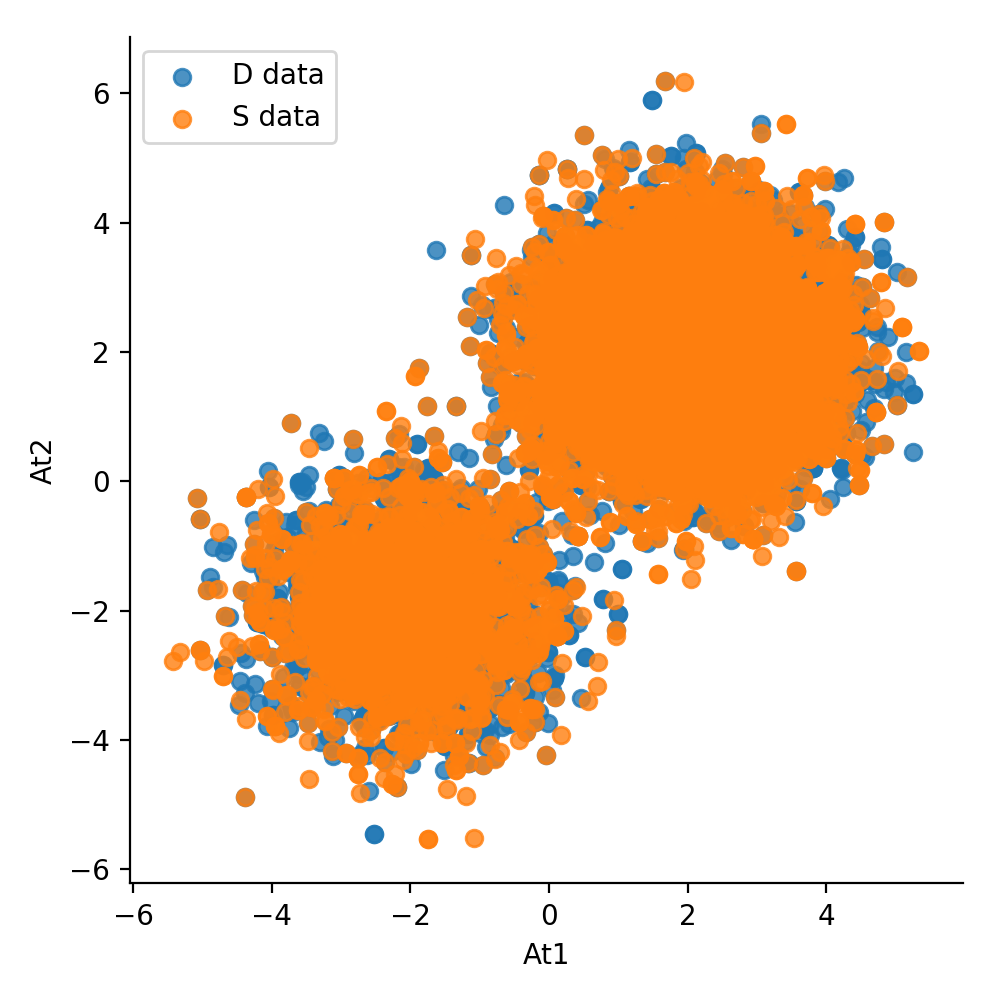}
  \caption{noisy S-D data (Labeling Noise)}
  \label{fig:5}
\end{subfigure}\hfil 
\begin{subfigure}{0.15\textwidth}
  \includegraphics[width=\linewidth]{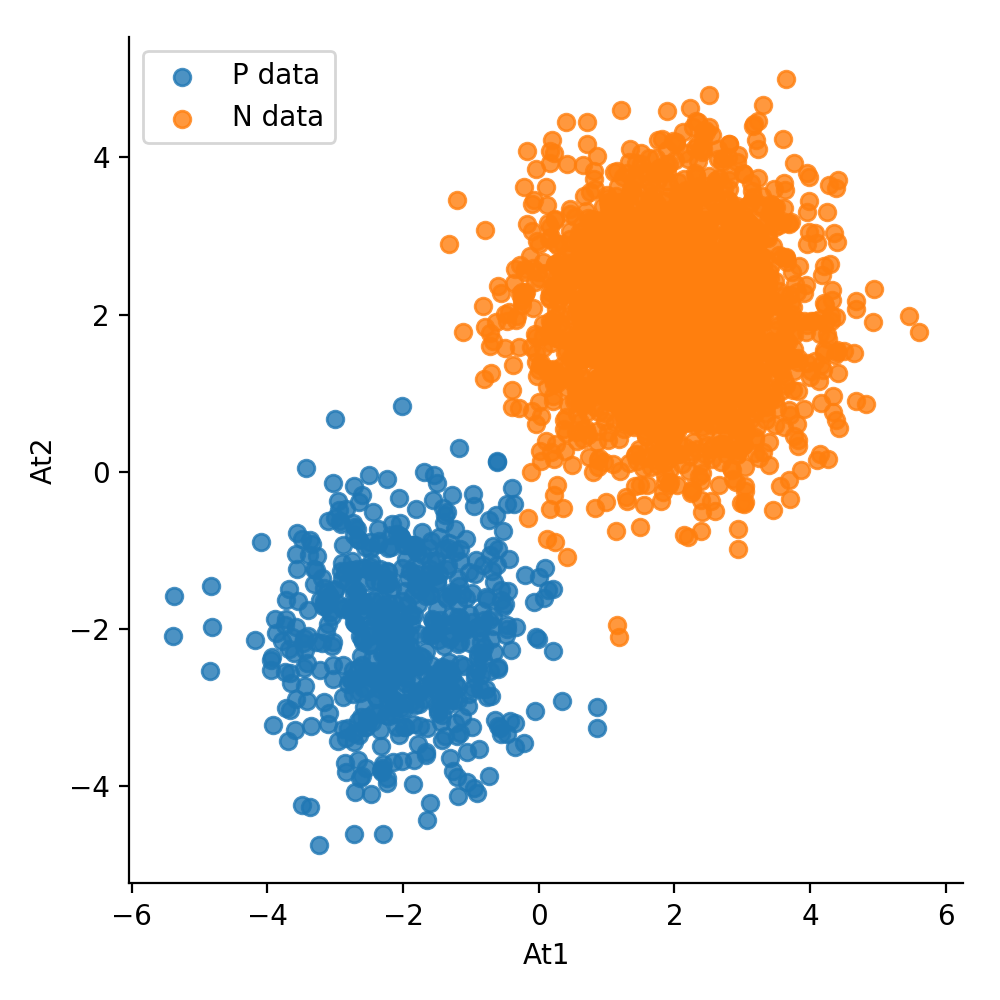}
  \caption{Weighted Classification: $99.67\%$ accuracy}
  \label{fig:6}
\end{subfigure}\hfil
\begin{subfigure}{0.15\textwidth}
  \includegraphics[width=\linewidth]{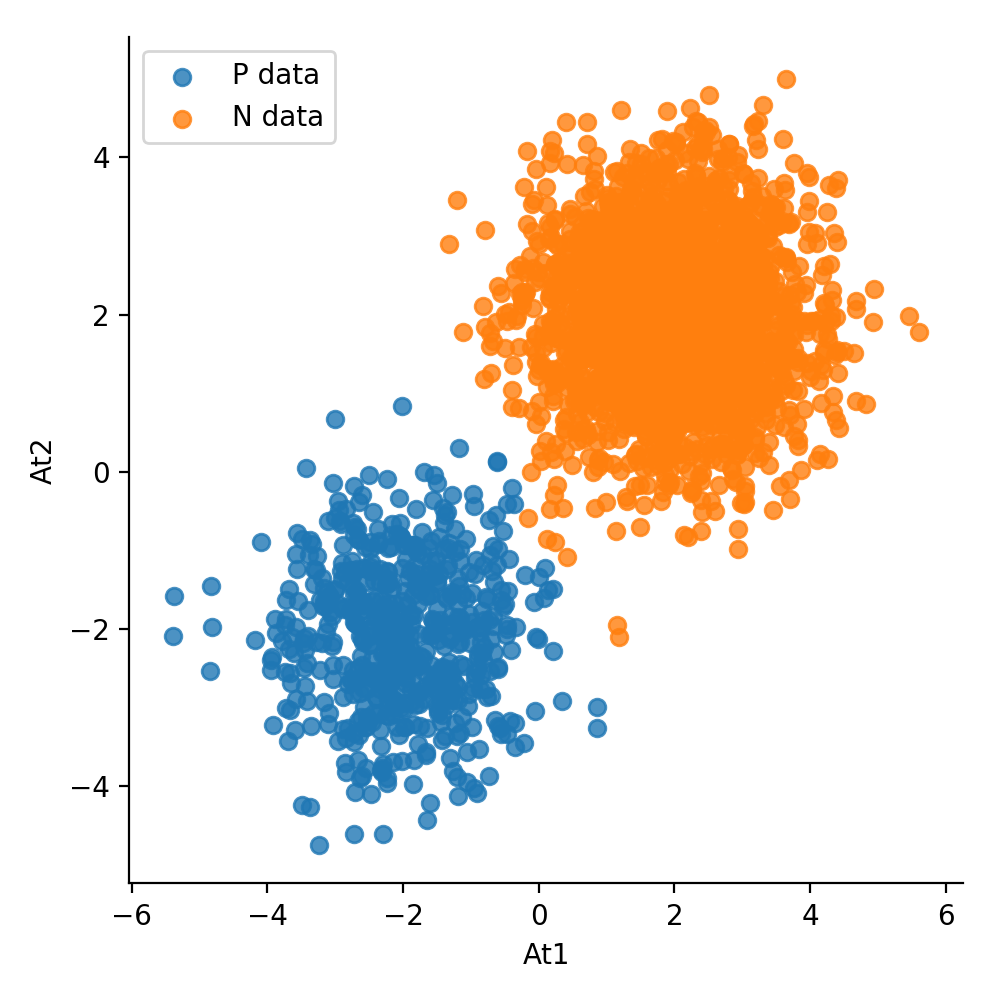}
  \caption{Loss Correction: $99.67\%$ accuracy}
  \label{fig:4}
\end{subfigure}

\caption{In this figure, we depict the underlying distribution with $\pi=0.2$ and the corresponding S and D data. The top row shows the test data accuracy of our proposed algorithms trained on the clean S-D training data. The bottom row shows the test data accuracy for the same test data but now training is done on noisy S-D data for each noise model with a high degree of noise (noise parameter set at $0.4$). We see that the proposed approaches perform well even under severe noise corruption.}
\label{fig:images}
\end{figure*}

\begin{table*}[t]
\centering
\caption{\textsc{Pairing Noise} : The best P-N column denotes the test accuracy after training on the standard class-labeled train dataset provided. $\protect d,\pi,N$ denote the feature dimension, class prior and the size of the entire class-labeled data respectively. clean S-D denotes test accuracy after training on clean S-D data generated from the train dataset. T-Loss indicates the test accuracy after training on S-D data with the loss correction approach (by the matrix $\protect T$) and SD-Loss denotes the non-corrected variant of \protect \cite{shimada2019classification}. Similarly, weighted and unweighted denotes the test accuracy after training on S-D data using weighted ERM and normal ERM respectively---note, they are identical for symmetric noise. KM denotes the KMeans baseline and KM-COP is the KMeans with constraints. Accuracies within $\protect 1\%$ of the best in each row are bolded. }\smallskip

\begin{tabular}{l|l|l|l|l|l|l|l|l|l}
\makecell{Dataset \\ $(d,\pi)$ \\ $N$} & 
best P-N &
clean S-D &
\makecell{Noise Rates \\ $(\rho_S,\rho_D)$} & 
T-Loss & 
SD-Loss &
Weighted &
Unweighted &
KM &
KM-COP  \\
\hline
\makecell{\textsc{diabetes} \\ (8,0.35)\\ 768 } &
\makecell{77} &
\makecell{77} &
\makecell{$(0.2,0.2)$ \\ $(0.1,0.2)$ \\ $(0.3,0.3)$ } &
\makecell{\textbf{76.57} \\ 74.95 \\ \textbf{75.52}} &
\makecell{75 \\ 75.52 \\ 73.95} &
\makecell{74.95 \\ \textbf{77.61} \\ 74.48} &
\makecell{74.95 \\ 76.04 \\ 74.48} &
\makecell{65.63}   &
\makecell{64.58 \\ 65.10 \\ 64.06} \\
\hline
\makecell{\textsc{adult} \\ (106,0.24)\\ 48842 } &
\makecell{83.09} &
\makecell{83.03} &
\makecell{$(0.2,0.2)$ \\ $(0.1,0.2)$ \\ $(0.3,0.3)$ } &
\makecell{\textbf{82.49} \\ 77.8 \\ \textbf{81.42}} &
\makecell{\textbf{82.22} \\ 76.26 \\ 80.26} &
\makecell{\textbf{82.35} \\ \textbf{82.41} \\ \textbf{81.10}} &
\makecell{\textbf{82.35} \\ 75.92 \\ \textbf{81.10}} &
\makecell{71.25}   &
\makecell{71.25 \\ 71.25 \\ 53.14} \\

\hline
\makecell{\textsc{cancer} \\ (30,0.37) \\569 } &
\makecell{97.2} &
\makecell{97.2} &
\makecell{$(0.2,0.2)$ \\ $(0.1,0.2)$ \\ $(0.3,0.3)$ } &
\makecell{\textbf{97.18} \\ \textbf{97.18} \\ \textbf{97.18}} &
\makecell{\textbf{96.47} \\ \textbf{97.18} \\ 95.07} &
\makecell{95.78 \\ 95.78 \\ 95.78} &
\makecell{95.78 \\ 95.78 \\ 95.78} &
\makecell{88.7}  &
\makecell{92.95 \\ 91.54 \\ 92.25} \\

\hline

\end{tabular}
\label{table1}
\end{table*}

\begin{table*}
\centering
\caption{\textsc{Labeling Noise}: The setup is same as Table \ref{table1} but now we use the labeling corruption noise model to generate the noisy S-D data.}

\begin{tabular}{l|l|l|l|l|l|l|l|l|l}
\makecell{Dataset \\ $(d,\pi)$\\ $N$ } & 
best P-N  &
clean S-D &
\makecell{Noise Rates \\ $(\rho_+,\rho_-)$} & 
T-Loss & 
SD-Loss &
Weighted &
Unweighted &
KM &
KM-COP  \\
\hline

\makecell{\textsc{ionosphere} \\ (34,0.64)\\351} &
\makecell{90.91} &
\makecell{90.91} &
\makecell{$(0.2,0.2)$ \\ $(0.1,0.2)$ \\ $(0.3,0.3)$ } &
\makecell{\textbf{88.67} \\ 85.24 \\ 87.5} &
\makecell{85.23 \\ 80.68 \\ 80.7} &
\makecell{86.4 \\ \textbf{90.91} \\ \textbf{88.64}} &
\makecell{86.4 \\ 85.23 \\ \textbf{88.64}} &
\makecell{70.45}  &
\makecell{71.59 \\ 71.59 \\ 71.59} \\

\hline
\makecell{\textsc{spambase} \\ (57,0.39)\\4601} &
\makecell{91.83} &
\makecell{89.74} &
\makecell{$(0.2,0.2)$ \\ $(0.1,0.2)$ \\ $(0.3,0.3)$ } &
\makecell{\textbf{87.56} \\ 83.74 \\ \textbf{85.304}} &
\makecell{83.22 \\ 84.15 \\ 75.65} &
\makecell{82.78 \\ \textbf{86.78} \\ 78.44} &
\makecell{82.78 \\ 85.56 \\ 78.44} &
\makecell{78.08} &
\makecell{78.96 \\ 79.13 \\78.61} \\
\hline

\makecell{\textsc{magic} \\ (10,0.65)\\19020} &
\makecell{84.12} &
\makecell{83.40} &
\makecell{$(0.2,0.2)$ \\ $(0.2,0.1)$ \\ $(0.3,0.3)$ } &
\makecell{80.06 \\ 73.27 \\ \textbf{79.39}} &
\makecell{81.13 \\ 73.61 \\ 78.42} &
\makecell{\textbf{82.21} \\ \textbf{81.67} \\ \textbf{79.50}} &
\makecell{\textbf{82.21} \\ 79.70 \\ \textbf{79.50}} &
\makecell{59.09} &
\makecell{63.28 \\ 66.03 \\ 62.34} \\
\hline
\end{tabular}

\label{table2}
\end{table*}

\section{Loss Correction Approach}
In this section, we present the first of our two proposed algorithms for learning from noisy pairwise data. In this method, we use the parameter values of the noise model to obtain an unbiased estimate of the loss function. In other words, we cast the loss of learning from noisy S-D data in terms of learning from standard class-labeled data and thus, we can do loss correction. In deriving the following expressions, we methodically account for all the different ways we can derive an S (or D instance) under each noise model (Figure \ref{fig:my_label}).
For the pairing noise model we obtain :
\begin{equation}
\label{pos1}
\begin{split}
P(Q=1|X) & = P(Y=1|X)[(1-\rho_S)\pi+\rho_D(1-\pi)] \\
    & +P(Y=-1|X)[\rho_D \pi+(1-\rho_S)(1-\pi)].
\end{split}
\end{equation}
On the other hand, for the labeling corruption noise model we obtain $ $:
\begin{equation}
\label{pos2}
\begin{split}
 P(Q=1|X) & =  P(Y=1|X)  [(1-\rho_+)\Tilde{\pi} + \rho_-(1-\Tilde{\pi})] \\
    & + P(Y=-1|X)[\rho_-\Tilde{\pi}+(1-\rho_+)(1-\Tilde{\pi})].
\end{split}
\end{equation}
We obtain similar equations for $P(Q=-1|X)$ under each noise model. It is noteworthy that the structure of the posterior probabilities is remarkably similar between the two noise models once we introduce the modified prior $\tilde{\pi}$, defined in \eqref{eq2}. In other words, the labeling noise model can be interpreted as first corrupting the ordinary class-labeled P (Positive)-N (Negative) data to get a noisy P-N data with the modified class prior and then mimicking the pairing noise model with noise rate $\rho_+,\rho_-$. Once we have expressed the posterior probabilities of the noisy S-D data in terms of the original class posteriors, we can adopt the technique of backward correction \cite{patrini2017making} to construct the corrected losses ---
Assume there exists coefficients $\alpha_1,\alpha_2,\beta_1,\beta_2$ such that $P(Q=1|x)= \alpha_1 P(y=1|x) + \alpha_2 P(y=-1|x)$ and   $P(Q=-1|x)= \beta_1 P(y=1|x) + \beta_2 P(y=-1|x)$, we can construct a new loss function such that the minimizer of the expected risk with this new loss function over the noisy S-D data is the same as the minimizer of the expected risk with the original loss function over $\mathcal{Z}$ (the test distribution). Let $l(t,y)$ be a loss function estimating $t$ by $y$ and let $\tilde{l} (t, Q)$ be the corresponding back-ward corrected loss function estimating $t$ by $Q$. Then we have $l(t,y)=T \tilde{l}(t,Q)$  where 
$T=
\begin{bmatrix} 
\alpha_1 & \alpha_2 \\
\beta_1 & \beta_2
\end{bmatrix}
$
and therefore, $\tilde{l}(t,Q)=T^{-1}l(t,y)$. We can use the modified loss to train our classifier $f \in F$(model class) on the noisy S-D data directly by ERM. Next, we state the generalization bounds for this approach. Here are some important notations:
\begin{itemize}
    
    \item
    $\hat{R}_{\tilde{l}}(f)=\frac{1}{n}\sum_{i=1}^n \tilde{l}(f(X_i),Q_i)$,
    \item 
    $ \hat{f} = \mathrm{argmin}_{f \in F} \hat{R}_{\tilde{l}}(f)$,
    \item
    $R_{\tilde{l},\mathcal{Z}_Q}(f)=E_{(X,Q) \sim \mathcal{Z}_Q}[\tilde{l}(f(X),Q)]$,
    \item
    $R_{l,\mathcal{Z}}(f)=E_{(X,Y) \sim \mathcal{Z}}[l(f(X),Y)]$.
\end{itemize}

Because $\tilde{l}$ is corrected appropriately, the empirical estimate of the risk converges to $R_{l,\mathcal{Z}}(f)$---by performing ERM on the noisy S-D data, the empirical risk converges to the true risk on the standard P-N data drawn from the underlying distribution $\mathcal{Z}$. Let $L_Q$ denote the Lipschitz constant of the loss $\tilde{l}$, $\mathcal{R}(F,N)$ denote the Rademacher complexity of the function class $F$ for $n$ noisy S-D instances.
\\
\textbf{Lemma 1} : With probability at least $1-\delta$,
\begin{equation}
    \max_{f \in F} |\hat{R}_{\tilde{l}}(f)-R_{\tilde{l},\mathcal{Z}_Q}(f)| \leq 2L_Q \mathcal{R}(F,n)+ \sqrt{\frac{\log(1/\delta)}{2n}}.
\end{equation}
\\
\textbf{Theorem 1} : With probability at least $1-\delta$,
\begin{equation}
R_{l,\mathcal{Z}}(\hat{f}) \leq \min_{f \in F} R_{l,\mathcal{Z}}(f)+4L_Q \mathcal{R}(F,n)+2\sqrt{\frac{\log(1/\delta)}{2n}}.
\end{equation}

 \textbf{Special Case of Clean S-D Learning} : When there is no noise $\rho_S$ = $\rho_D$ = $0$ or $\rho_+$ = $\rho_-$ = $0$, for both the noise models, 
\begin{equation}
\begin{split}
T = 
\begin{bmatrix} 
\pi & 1-\pi \\
1-\pi & \pi 
\end{bmatrix} &
\implies T^{-1} = 
\begin{bmatrix} 
\frac{\pi}{(2\pi-1)} & \frac{-(1-\pi)}{(2\pi-1)} \\
\frac{-(1-\pi)}{(2\pi-1)} & \frac{\pi}{(2\pi-1)}
\end{bmatrix}
\end{split}
.
\end{equation}
We see this matches the loss function derived for clean S-D learning in \cite{shimada2019classification} (on setting $X'$ as $X$ and replacing $\pi_S E_{X_S}[1] = \pi_D E_{X_D}[1] = \frac{1}{n}$) : 
\begin{equation}
\begin{split}
&\hat{R}_{\tilde{l}}(f) =\frac{1}{n} \sum_{i=1}^n [\mathcal{L}(f(X_i),Q_i)], \\
   \mathrm{where \quad} &\mathcal{L}(z,t)=\frac{\pi}{2\pi-1}l(z,t)-\frac{1-\pi}{2\pi-1}l(z,-t).
\end{split}
\end{equation}
Our analysis through the lens of loss correction helps provide the generalization bound for clean S-D learning as a special case of noisy S-D learning. 
The Lipschitz constant for the corrected loss in the clean S-D case is $L_Q = \frac{L}{|2\pi-1|}$, where $L$ is the Lipschitz constant for $l$. The generalization bound for noise-free S-D learning is : With probability at least $1-\delta$,
\begin{equation}
R_{l,\mathcal{Z}}(\hat{f}) \leq \min_{f \in F} R_{l,\mathcal{Z}}(f)+\frac{4L}{|2\pi-1|} \mathcal{R}(F,n)+2\sqrt{\frac{\log(1/\delta)}{2n}}.
\end{equation}
\\
\textbf{Optimization} : While we have a generalization guarantee, efficient optimization is a concern especially because the corrected loss $\tilde{l}(\cdot,\cdot)$ may not be convex. We present a condition which will guarantee the corrected loss to be convex.
\\
\textbf{Theorem 2} : If $l(t,y)$ is convex and twice differentiable almost everywhere in $t$ (for each $y$) and also satisfies : 
\begin{itemize}
    \item 
$\forall t \in Q, \quad l''(t,y)=l''(t,-y)$,
\item
$\mathrm{sign}(\alpha_1-\beta_1)=\mathrm{sign}(\beta_2-\alpha_2)$, where $\alpha_1,\alpha_2,\beta_1,\beta_2$ are elements of $T$.
\end{itemize}
then $\tilde{l}(t,y)$ is convex in $t$.
The first condition is satisfied by several common losses such as squared loss ($l(y,t)=(1-ty)^2$) and logistic loss ($l(t,y)=\log(1+\exp(-ty))$). 
The second condition depends on the noise rates and the prior. We can simplify this for each noise model to : 
\begin{equation}
\begin{split}
\frac{1-2\rho_S}{1-2\rho_D} & \in \left[\frac{1-\pi}{\pi}, \frac{\pi}{1-\pi}\right] \mathrm{\quad or  \quad} \left[\frac{\pi}{1-\pi}, \frac{1-\pi}{\pi}\right],
\\
    \frac{1-2\rho_+}{1-2\rho_-} & \in \left[\frac{1-\Tilde{\pi}}{\Tilde{\pi}}, \frac{\Tilde{\pi}}{1-\Tilde{\pi}}\right] \mathrm{\quad or  \quad} \left[\frac{\Tilde{\pi}}{1-\Tilde{\pi}},\frac{1-\Tilde{\pi}}{\Tilde{\pi}}\right].
    \end{split}
\end{equation}
where either the first bound or the second bound applies depending on whether the class prior is greater or less than $0.5$ respectively. For all cases of noise-free or symmetric-noise S-D learning, any noise rates will satisfy this condition and thus, we can always perform efficient optimization. 

\begin{figure*}
    \centering 
\begin{subfigure}{0.23\textwidth}
  \includegraphics[width=\linewidth]{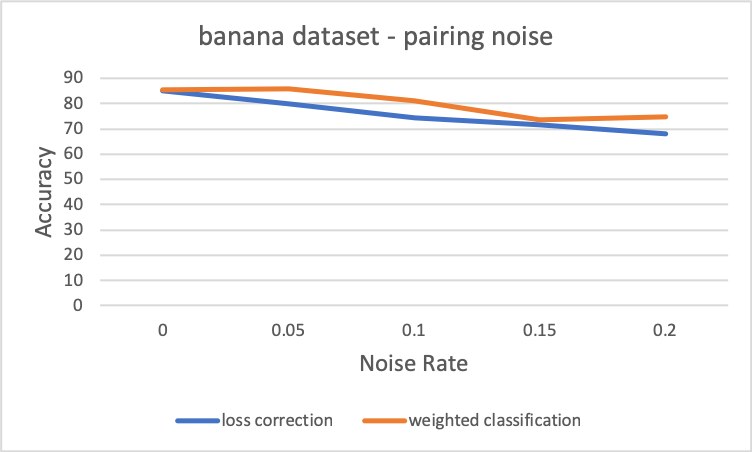}
  \label{fig:1}
\end{subfigure}\hfil 
\begin{subfigure}{0.23\textwidth}
  \includegraphics[width=\linewidth]{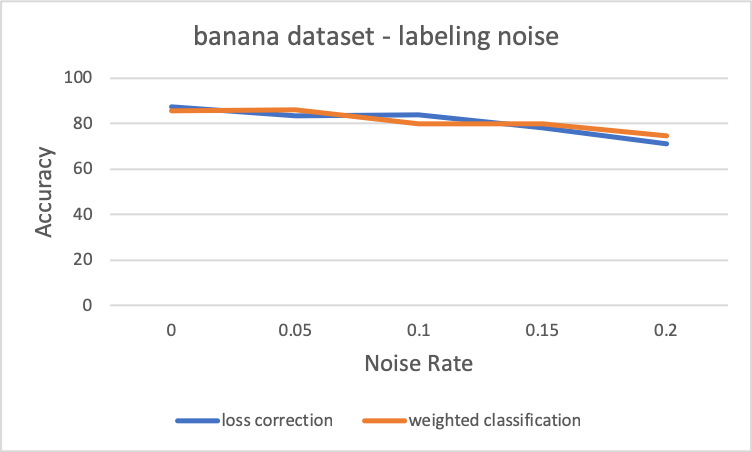}
  \label{fig:2}
\end{subfigure}\hfil 
\begin{subfigure}{0.23\textwidth}
  \includegraphics[width=\linewidth]{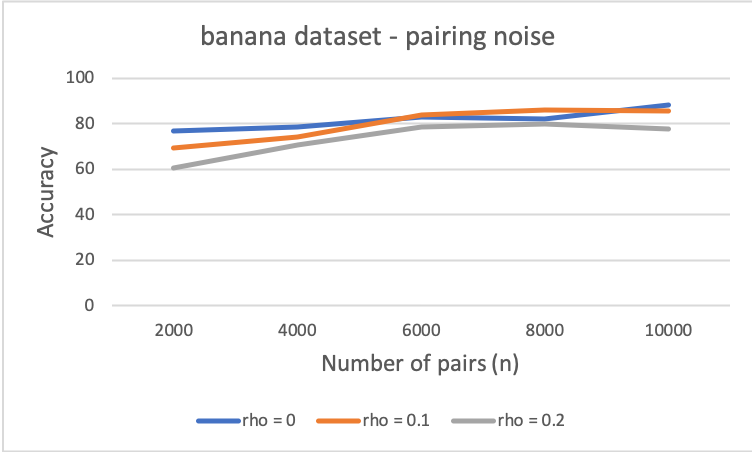}
  \label{fig:3}
\end{subfigure}
\begin{subfigure}{0.23\textwidth}
  \includegraphics[width=\linewidth]{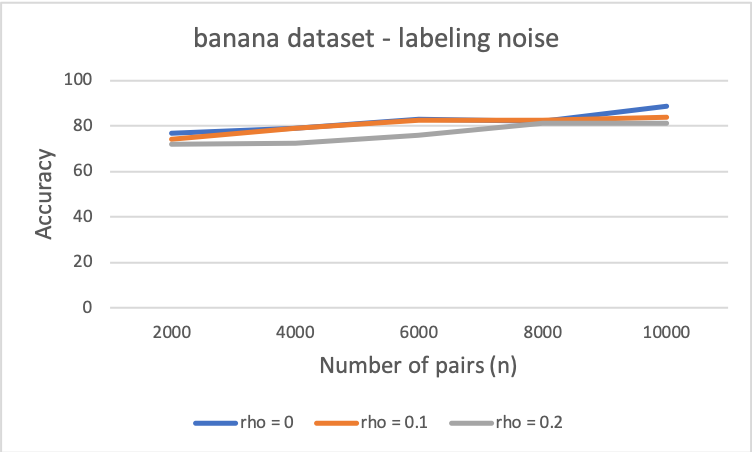}
  \label{fig:2}
\end{subfigure}\hfil 

\caption{The first two images depict the gradual decrease in classification accuracy of the learned classifier (from either algorithm) on the test dataset as the noise rate in the noisy S-D training data increases, for each noise model. The last two images depict the gradual increase in classification accuracy of the learned classifier (from the weighted classification method) on the test dataset as the number of noisy S-D training samples increases, for different noise rates in each noise model. The accuracy achieved from training on clean P-N training data provided by the banana dataset is $90.8\%$ }
\label{fig:imagesb}
\end{figure*}

\section{Weighted Classification Approach}
Now we develop our second algorithm for dealing with noisy S and D data. One key issue that we investigate here is how the Bayes classifier learned from noisy S-D data relates to the traditional Bayes classifier. \\
\textbf{Lemma 2} : Denoting $P(Q=1|x) = \eta_Q(x)$ and $P(Y=1|x)= \eta(x)$, the Bayes classifier under the noisy S-D distribution denoted by $\Tilde{f}^* = \mathrm{argmin}_{f \in F} E_{(X,Q) \sim \mathcal{Z}_Q}[{1}_{\lbrace \mathrm{sign}(f(X))\neq Q \rbrace }]$ is given by 
\begin{equation}
\label{fin}
\tilde{f}^*(x)=\mathrm{sign}\left(\eta_Q(x)-\frac{1}{2}\right)=\mathrm{sign}\left(\eta(x)-T\right)
\end{equation}
For the pairwise corruption case,
$$
T= \frac{\frac{1}{2}-[(1-\rho_S)(1-\pi)+\rho_D\pi]}{(1-\rho_S-\rho_D)(2\pi-1)}.
$$
For the label corruption case, threshold $T$ is :
$$
\frac{\frac{1}{2}-\pi(\rho_{+}+\rho_{-}-\rho_{+}\rho_{-})-(1-\pi)(\rho_{+}^2+(1-\rho_{-})^2)}{(1-\rho_{+}-\rho_{-})[\pi(1-2\rho_{+})-(1-\pi)(1-2\rho_{-})]}.
$$
These expressions can be derived by using \eqref{pos1} and \eqref{pos2} in \eqref{fin}.
They give us an important insight---the Bayes classifier for the noisy S-D learning uses a different threshold from $\frac{1}{2}$ while the traditional Bayes classifier has $\eta(x)$ thresholded at $\frac{1}{2}$. Towards designing an algorithm we note that we can also obtain this Bayes classifier by minimizing the weighted 0-1 risk. Let 
$$
U_\alpha(t,y)=(1-\alpha)1_{\lbrace y=1 \rbrace }1_{\lbrace t \leq 0 \rbrace}+\alpha1_{\lbrace y=-1 \rbrace}1_{\lbrace t > 0 \rbrace}.
$$
The following lemma from \cite{scott2012calibrated} is crucial in connecting the Bayes classifier threshold with the weight in weighted 0-1 classification.
\\
\textbf{Lemma 3} : Denote the $U_\alpha$ risk under distribution $\mathcal{Z}$ as
$$
R_{\alpha,D}(f)=E_{(x,y) \sim \mathcal{Z}}[U_\alpha(f(x),y)].
$$
Then $f^*_\alpha(x)=\mathrm{sign}(\eta(x)-\alpha)$ minimizes $R_{\alpha,\mathcal{Z}}(f)$.\\

We now show that there exists a choice of weight $\alpha$ such that the weighted risk under the noisy S-D distribution is linearly related to the ordinary risk under distribution $\mathcal{Z}$. \\

\textbf{Theorem 3} : There exist constants $\alpha$ and $A$ and a function $B(X)$ (that only depends on $X$ but not on $f$) such that 
$$
R_{\alpha,\mathcal{Z}_Q}(f)=A R_{\mathcal{Z}}(f)+E_X[B(X)].
$$
For the pairing corruption case:
\begin{equation}
\begin{split}
    \alpha & =\frac{1-\rho_S+\rho_D}{2},
    \\
    A & =\frac{1-\rho_S-\rho_D}{2}(2\pi-1).
    \end{split}
\end{equation}

For the label corruption case :
\begin{equation}
\begin{split}
    \alpha & =\pi(1-\rho_{+}+\rho_{+}^2-\rho_{+}\rho_{-})
-\frac{1}{2}(1-\rho_{+}-\rho_{-}) \\
    & +(1-\pi)(1-\rho_{-}+\rho_{-}^2-\rho_{+}\rho_{-}),
    \\
    A & =\frac{(1-\rho_{+}-\rho_{-})}{2}[\pi(1-2\rho_{+})-(1-\pi)(1-2\rho_{-})].
    \end{split}
\end{equation}
\textbf{Remark 1}: The $\alpha$-weighted Bayes optimal classifier under the noisy S-D distribution coincides with the Bayes classifier of the 0-1 loss under the clean ordinary distribution.
\\
\textbf{Generalization Bound} : For the ease of optimization, we will use a surrogate loss instead of 0-1 to do weighted ERM. Any surrogate loss can be used as long as it can be decomposed as $l(t,Q)=1_{\lbrace Q=1 \rbrace}l_1(t)+1_{\lbrace Q=-1 \rbrace}l_{-1}(t)$. The square, hinge, logistic and most common losses can be expressed in this form.
Let $\hat{f}_\alpha$ denote the minimizer of the empirical risk using the weighted surrogate loss.
\begin{equation}
    \hat{f}_\alpha = \mathrm{argmin}_{f \in F} \frac{1}{n} \sum_{i=1}^n l_\alpha(f(x),Q).
\end{equation}
\textbf{Theorem}: If $l_\alpha$ is an $\alpha$-weighted margin loss of the form : $l_\alpha(t,Q)=(1-\alpha)1_{\lbrace Q=-1 \rbrace}l(t)+\alpha1_{\lbrace Q=1 \rbrace}l(-t)$ and $l$ is classification calibrated and $L$-Lipschitz, then for the above choices of $\alpha$ and $A$, there exists a non-decreasing function $\xi_{l_{\alpha}}$ with $\xi_{l_{\alpha}}(0)=0$ such that the following bound holds with probability at least $1-\delta$:
\begin{equation}
\begin{split}
R_{\mathcal{Z}}(\hat{f}_{\alpha}) -R^* \leq 
A^{-1} \xi_{l_{\alpha}}\bigg(\min_{f \in F} R_{\alpha,\mathcal{Z}_Q}(f)-
\\
\min_{f} R_{\alpha,\mathcal{Z}_Q}(f)+4L \mathcal{R}(F,n)+2\sqrt{\frac{\log(1/\delta)}{2n}} )\bigg).
\end{split}
\end{equation}
where $R^*$ denotes the the corresponding Bayes risk under $\mathcal{Z}$.\\
\textbf{Remark 2}: For a fixed Lipschitz constant $L$, as $A$ decreases we get a weaker generalization bound. For the pairing noise model, its easy to see that as noise rates increase, $A$ decreases but the relationship is more complicated for the labeling noise model. When the noise is symmetric ($\rho_+=\rho_-=\rho$), $A=\frac{(1-2\rho)^2(2\pi-1)}{2}$. In this case, again we observe as $\rho$ increases, $A$ decreases and we get a weaker bound.
\\
\textbf{Remark 3}: When $\rho_S=\rho_D$ or $\rho_+=\rho_-$, we see that the optimal Bayes classifier for the (noisy) S-D learning problem is the same as the Bayes classifier for the standard class-labeled binary classification task under distribution $\mathcal{Z}$. \\ 
\section{Estimation of Prior and Noise Parameters}
We briefly discuss the parameters we need to know or estimate to apply each method for each noise model.\\
\textbf{Loss Correction approach}: The noise rate parameters can be tuned by cross-validation on the noisy S-D data.
We also need to estimate the class prior to construct the loss correction matrix $T$, under both noise models. The class prior can be estimated as follows :
\begin{itemize}
    \item For the pairing corruption noise model : \\
$\frac{n_S}{n_D}=\frac{(1-\rho_S)(\pi^2+(1-\pi)^2)+2\rho_D\pi(1-\pi)}{\rho_S(\pi^2+(1-\pi)^2)+2(1-\rho_D)\pi(1-\pi)}$

\item For the labeling corruption noise model : \\
$\frac{n_S}{n_D}=\frac{(1-\rho_+)(\pi\Tilde{\pi}+(1-\pi)(1-\Tilde{\pi}))+\rho_-(\pi(1-\Tilde{\pi})+\Tilde{\pi}(1-\pi))}{\rho_+(\pi\Tilde{\pi}+(1-\pi)(1-\Tilde{\pi}))+(1-\rho_-)(\pi(1-\Tilde{\pi})+\Tilde{\pi}(1-\pi))}$ 
\end{itemize}
From each of the above equations we can solve for the class prior $\pi$. The above equations can be derived from \eqref{pos1} and \eqref{pos2} by marginalizing out $X$ and using $\frac{n_S}{n_D}=\frac{P(Q=1)}{P(Q=-1)}$.

 \textbf{Method of Weighted Classification}: The class prior only appears in the weight $\alpha$ in the labeling corruption model. In the pairing corruption model, knowledge of the class prior is not needed to calculate $\alpha$. However, since we just have one parameter $\alpha$ for the optimization problem, in practice we obtain $\alpha$ directly by cross-validation under both noise models.

\section{Experiments}
We empirically verify that the proposed algorithms is able to learn a classifier for the underlying distribution $\mathcal{Z}$ from only noisy similar and dissimilar training data.
All experiments are repeated $3$ times on random train-test splits of $75$:$25$ and the average accuracies are shown.
We conduct experiments on two noise models independently.
In the learning phase, the noise parameters and the weight $\alpha$ is tuned by cross-validation.
Evaluation is done on the clean class-labeled test dataset using standard classification accuracy as evaluation metric which is averaged over the test datasets to reduce variance across the corruption in the training data.
We use a multi-layer perceptron (MLP) with two hidden layers of $100$ neurons, ReLU activation and a single logistic sigmoid output, as our model architecture for all our experiments except the synthetic generated data, where we use a logistic linear model. We use the squared loss throughout : $l(t,y)=(t-y)^2$ and use Stochastic Gradient Descent (SGD) with a learning rate of $0.001$ and momentum of $0.9$.

\subsection{Synthetic Data}

We generate a two-dimensional dataset of points sampled from a normal distribution $N((2,2), \mathcal{I}_2)$ as one class and  $N((-2,-2), \mathcal{I}_2)$ as the other.
Here $\mathcal{I}_2$ denotes the two-dimensional identity matrix indicating that they are isotropic.
Note that the dataset generated in this manner may be non-separable.
In Figure~\ref{fig:images}, we show the data distribution, the test data, the noisy S-D data and the performance of the classifier learned from the noisy S-D data on the test data.

We use a non-separable benchmark ``banana'' dataset which has two-dimensional attributes and two classes. We perform two kinds of experiments. In the first experiment, for a given noise model, for different settings of symmetric noise parameters ($\rho_S=\rho_D$ and $\rho_+=\rho_-$) we plot the variation of clean test accuracy with the number of noisy S-D pairs ($n$) sampled for training.
For this experiment setting, we show the results for the weighted classification algorithm in Figure \ref{fig:imagesb}.
Since the Bayes classifier under the symmetric noise is identical to that of noise-free case under both the noise models,
we see that the accuracy improves as we get a better approximation of the Bayes classifier as we have more and more S-D data-points in training.
Note that the number of original training points in the dataset is fixed---what changes is only the number of S-D points we sample from them.
In the second experiment, for each noise model, for a fixed $n$ we show the gradual degradation of performance of the proposed algorithms (loss correction approach as well as the weighted classification approach) with increasing symmetric noise rates.
These experiments confirm that higher noise hurts classification accuracy and having more pairwise samples help it.

\subsection{Real World datasets}
We further conduct experiments on several benchmark datasets from the UCI classification tasks.\footnote{Available at \url{https://archive.ics.uci.edu/ml/datasets.php}.}
All tasks are binary classification tasks of varying dimensions, class priors, and sample sizes.
We compare the performance of our proposed approaches against two kinds of baselines.

 Supervised Baseline : the noise-free versions of the proposed approaches are used, i.e, for loss correction approach we assume clean S-D classification risk formulation~\cite{shimada2019classification} and for weighted classification approach we assume clean S-D case and threshold at $\frac{1}{2}$. This baseline is simple but since there are no existing supervised algorithms that can handle S-D data except \cite{shimada2019classification} and \cite{bao2018classification}, and none for the noisy S-D supervision case, that is the only baseline to compare against that makes use of the whole data and supervision provided. 

 Unsupervised Baseline :  We also compare against unsupervised clustering and semi-supervised clustering based methods. For unsupervised clustering, we ignore all pairwise information and apply KMeans~\cite{macqueen1967some} with, number of clusters as $2$, directly on the noisy S-D datapoints and use the obtained clusters to classify the test data. We further use constrained KMeans clustering~\cite{wagstaff2001constrained}, where we treat the S-D pairs as constraints to supervise the clustering of the S-D data pooled together. These two baselines are shown in Tables~\ref{table1} and~\ref{table2}. 

 From Tables~\ref{table1} and~\ref{table2}, we observe that for both noise models and for every noise rate, our proposed approaches outperform the baselines. Further, we see that the clean S-D performances match the best P-N performance which empirically verifies the optimal classifiers for learning from noise-free standard P-N and pairwise S-D data coincide . 

\section{Conclusion and Future Work}
In this paper we investigated a novel setting---learning from noisy pairwise labels under two different noise models. We showed the connections of this problems to standard class-labeled binary classification, proposed two algorithms and analyzed their generalization bounds. We empirically showed that they outperform supervised and unsupervised baselines. For future work, it is worthwhile to investigate more complicated noise models such as instance-dependent noise~\cite{menon2016learning} in this setting.

\bibliographystyle{named}
\bibliography{ijcai20}

\begin{thebibliography}{}

\bibitem[\protect\citeauthoryear{Bao \bgroup \em et al.\egroup
  }{2018}]{bao2018classification}
Han Bao, Gang Niu, and Masashi Sugiyama.
\newblock Classification from pairwise similarity and unlabeled data.
\newblock In {\em International Conference on Machine Learning}, pages
  461--470, 2018.

\bibitem[\protect\citeauthoryear{Basu \bgroup \em et al.\egroup
  }{2008}]{basu2008constrained}
Sugato Basu, Ian Davidson, and Kiri Wagstaff.
\newblock {\em Constrained clustering: Advances in algorithms, theory, and
  applications}.
\newblock CRC Press, 2008.

\bibitem[\protect\citeauthoryear{Elkan}{2001}]{elkan2001foundations}
Charles Elkan.
\newblock The foundations of cost-sensitive learning.
\newblock In {\em International joint conference on artificial intelligence},
  volume~17, pages 973--978. Lawrence Erlbaum Associates Ltd, 2001.

\bibitem[\protect\citeauthoryear{Eric \bgroup \em et al.\egroup
  }{2008}]{eric2008active}
Brochu Eric, Nando~D Freitas, and Abhijeet Ghosh.
\newblock Active preference learning with discrete choice data.
\newblock In {\em Advances in neural information processing systems}, pages
  409--416, 2008.

\bibitem[\protect\citeauthoryear{Fisher}{1993}]{fisher1993social}
Robert~J Fisher.
\newblock Social desirability bias and the validity of indirect questioning.
\newblock {\em Journal of consumer research}, 20(2):303--315, 1993.

\bibitem[\protect\citeauthoryear{F{\"u}rnkranz and
  H{\"u}llermeier}{2010}]{furnkranz2010preference}
Johannes F{\"u}rnkranz and Eyke H{\"u}llermeier.
\newblock {\em Preference learning}.
\newblock Springer, 2010.

\bibitem[\protect\citeauthoryear{Han \bgroup \em et al.\egroup
  }{2018}]{han2018co}
Bo~Han, Quanming Yao, Xingrui Yu, Gang Niu, Miao Xu, Weihua Hu, Ivor Tsang, and
  Masashi Sugiyama.
\newblock Co-teaching: Robust training of deep neural networks with extremely
  noisy labels.
\newblock In {\em Advances in neural information processing systems}, pages
  8527--8537, 2018.

\bibitem[\protect\citeauthoryear{Jamieson and Nowak}{2011}]{jamieson2011active}
Kevin~G Jamieson and Robert Nowak.
\newblock Active ranking using pairwise comparisons.
\newblock In {\em Advances in Neural Information Processing Systems}, pages
  2240--2248, 2011.

\bibitem[\protect\citeauthoryear{Jamieson \bgroup \em et al.\egroup
  }{2012}]{jamieson2012query}
Kevin~G Jamieson, Robert Nowak, and Ben Recht.
\newblock Query complexity of derivative-free optimization.
\newblock In {\em Advances in Neural Information Processing Systems}, pages
  2672--2680, 2012.

\bibitem[\protect\citeauthoryear{Jiang \bgroup \em et al.\egroup
  }{2017}]{jiang2017mentornet}
Lu~Jiang, Zhengyuan Zhou, Thomas Leung, Li-Jia Li, and Li~Fei-Fei.
\newblock Mentornet: Regularizing very deep neural networks on corrupted
  labels.
\newblock {\em arXiv preprint arXiv:1712.05055}, 4, 2017.

\bibitem[\protect\citeauthoryear{MacQueen and others}{1967}]{macqueen1967some}
James MacQueen et~al.
\newblock Some methods for classification and analysis of multivariate
  observations.
\newblock In {\em Proceedings of the fifth Berkeley symposium on mathematical
  statistics and probability}, volume~1, pages 281--297. Oakland, CA, USA,
  1967.

\bibitem[\protect\citeauthoryear{Menon \bgroup \em et al.\egroup
  }{2016}]{menon2016learning}
Aditya~Krishna Menon, Brendan Van~Rooyen, and Nagarajan Natarajan.
\newblock Learning from binary labels with instance-dependent corruption.
\newblock {\em arXiv preprint arXiv:1605.00751}, 2016.

\bibitem[\protect\citeauthoryear{Natarajan \bgroup \em et al.\egroup
  }{2013}]{natarajan2013learning}
Nagarajan Natarajan, Inderjit~S Dhillon, Pradeep~K Ravikumar, and Ambuj Tewari.
\newblock Learning with noisy labels.
\newblock In {\em Advances in neural information processing systems}, pages
  1196--1204, 2013.

\bibitem[\protect\citeauthoryear{Patrini \bgroup \em et al.\egroup
  }{2017}]{patrini2017making}
Giorgio Patrini, Alessandro Rozza, Aditya Krishna~Menon, Richard Nock, and
  Lizhen Qu.
\newblock Making deep neural networks robust to label noise: A loss correction
  approach.
\newblock In {\em Proceedings of the IEEE Conference on Computer Vision and
  Pattern Recognition}, pages 1944--1952, 2017.

\bibitem[\protect\citeauthoryear{Saaty}{1990}]{saaty1990decision}
Thomas~L Saaty.
\newblock {\em Decision making for leaders: the analytic hierarchy process for
  decisions in a complex world}.
\newblock RWS publications, 1990.

\bibitem[\protect\citeauthoryear{Scott and others}{2012}]{scott2012calibrated}
Clayton Scott et~al.
\newblock Calibrated asymmetric surrogate losses.
\newblock {\em Electronic Journal of Statistics}, 6:958--992, 2012.

\bibitem[\protect\citeauthoryear{Shimada \bgroup \em et al.\egroup
  }{2019}]{shimada2019classification}
Takuya Shimada, Han Bao, Issei Sato, and Masashi Sugiyama.
\newblock Classification from pairwise similarities/dissimilarities and
  unlabeled data via empirical risk minimization.
\newblock {\em arXiv preprint arXiv:1904.11717}, 2019.

\bibitem[\protect\citeauthoryear{Thurstone}{1927}]{thurstone1927law}
Louis~L Thurstone.
\newblock A law of comparative judgment.
\newblock {\em Psychological review}, 34(4):273, 1927.

\bibitem[\protect\citeauthoryear{Wagstaff \bgroup \em et al.\egroup
  }{2001}]{wagstaff2001constrained}
Kiri Wagstaff, Claire Cardie, Seth Rogers, Stefan Schr{\"o}dl, et~al.
\newblock Constrained k-means clustering with background knowledge.
\newblock In {\em Icml}, volume~1, pages 577--584, 2001.

\end{thebibliography}

\end{document}